\RequirePackage{fix-cm}
\documentclass[11pt]{article}          
\usepackage{graphicx}
\usepackage{algorithm}
\usepackage{algorithmic}
\usepackage{epsfig}
\usepackage{blindtext}
\usepackage{pgfplots}
\usepackage{tikz}
\usepgflibrary{plotmarks} 
\usepgflibrary[plotmarks] 
\usetikzlibrary{plotmarks} 
\usetikzlibrary[plotmarks] 
\usetikzlibrary{shapes,arrows}

\tikzstyle{decision} = [diamond, draw,
    text width=5em, text badly centered, node distance=2cm, inner sep=0pt]
\tikzstyle{block} = [rectangle, draw,
    text width=8em, text centered, rounded corners, minimum height=1em]
\tikzstyle{line} = [draw, -latex']
\tikzstyle{cloud} = [draw, ellipse, node distance=1cm,
    minimum height=1em]

\usetikzlibrary{chains,fit,shapes}
\begin{document}
\title{Random-Key Cuckoo Search for the Travelling Salesman Problem}
\author{Aziz Ouaarab,
        Bela\"{i}d Ahiod, 	
        Xin-She Yang
\\ \\ \hline
    A. Ouaarab and B. Ahiod \\
              LRIT, Associated Unit to the CNRST(URAC) no 29,\\
			  Mohammed V-Agdal University, B.P. 1014 Rabat, Morocco \\
			  \and
			  Xin-She Yang \\
			  School of Science and Technology, Middlesex University,\\
			  The burroughs, London NW4 4BT, UK\\
\hline
}

\date{{\bf Citation detail}: A. Ouaarab, B. Ahiod, X. S. Yang, Random-key cuckoo search for the travelling salesman problem,
{\it Soft Computing}, 19(4), 1099-1106(2015). }

\maketitle

\begin{abstract}
\label{abstract}
Combinatorial optimization problems are typically NP-hard, and thus very challenging to solve.
In this paper, we present the random key cuckoo search (RKCS) algorithm for solving the famous Travelling Salesman
Problem (TSP). We used a simplified random-key encoding scheme
to pass from a continuous space (real numbers) to a combinatorial space. We also consider the displacement of
a solution in both spaces using L\'evy flights. The performance of the proposed RKCS is tested against a set of benchmarks of
symmetric TSP from the well-known TSPLIB library. The results of the tests show that RKCS is superior to
	some other metaheuristic algorithms.

\end{abstract}

\section{Introduction}
\label{intro}
Many combinatorial optimization problems are NP-hard, and thus very challenging to solve. In fact, they cannot be solved efficiently by any
known algorithm in a practically short time scale when the size of the problem is moderately large~\cite{hoch96}.
The main difficulty arises with the number of combinations which increases
exponentially with the size of the problem. Searching for every possible combination is extremely
computationally expansive and unrealistic. An example of these problems is the travelling salesman problem~\cite{lawler85} in which a salesperson has to visit a list of cities exactly once, and returning to the departure city, with the aim of minimizing the total travelled distance or the overall cost of the trip.

 Despite the challenges, TSP remains one of the most widely studied problems in combinatorial optimization. It is often used for testing optimization algorithms. Problems such as TSP do not have an efficient algorithm to solve them. It is practically very difficult to get a solution of optimal quality and in a reduced runtime simultaneously. This requires some heuristic algorithms that can find good (not necessarily optimal) solutions in a good runtime by trial and error. Approximate algorithms such as metaheuristics~\cite{blum03} are actually the best choice to solve many combinatorial optimization problems. They are characterized by their simplicity and flexibility while demonstrating  remarkable effectiveness. Metaheuristics are usually simple to implement; however, they are often capable to solve complex problems and can thus be adapted to solve problems with diverse objective function properties, either continuous, discrete or mixed, including  many real-world optimization problems, from engineering to artificial intelligence~\cite{yang2010e}.

  In essence, metaheuristics are optimization algorithms that adopt some strategies to explore and exploit a given solution space with the aim to find the best solution. They balance their search concentration between some promising regions and the all space regions. They generally begin with a set of initial solutions, and then, examine step by step a sequence of solutions to reach (hopefully) the optimal solution of the problem.

Some issues may arise when solving a combinatorial optimization problem with
a metaheuristic, and a key issue is how to define neighbourhood solutions for such problems.
Several metaheuristics are designed in principle for continuous optimization problems. So,
the question is how to treat combinatorial problems properly without losing the good performance of these metaheuristics.
In this paper, we propose the random-key cuckoo search (RKCS) algorithm using the
random-key encoding scheme to represent a  position, found by the cuckoo search algorithm, in a combinatorial space.

TSP is solved with random keys by various metaheuristics~\cite{sama07h,chen11}.
This work presents a novel approach using Cuckoo Search algorithm (CS)\cite{yang09}, based on
random keys~\cite{bean1994genetic}, with a simple local search procedure to solve TSP.
CS is a nature-inspired metaheuristic algorithm which was developed by Yang and Deb
in 2009 to solve continuous optimization problems. With this approach, we aim to formulate
the transition between a continuous search space and a combinatorial search space without passing through traditional adaptation operators that
may affect the performance of the algorithm, and to ensure a direct interpretation of various operators used by metaheuristics in continuous search space.

The rest of this paper is organized as follows: Sect.~2, first,
introduces the standard cuckoo search. Section~3 introduces briefly the TSP. Section~4 presents the random-key encoding scheme,
while Sect.~5 describes the discrete CS to solve symmetric TSP using Random key. Then, Sect.~6 presents results of numerical experiments on a set of benchmarks of
symmetric TSP from the TSPLIB library~\cite{reinelt95}. Finally, Sect.~7 concludes with some discussions and future directions.

\section{Cuckoo Search Algorithm}
\label{sec:2}
Some cuckoo species can have the so-called brood parasitism as an aggressive reproduction strategy.
This most studied and discussed feature is that cuckoos lay eggs in a previously observed nest of another species
to let host birds hatch and brood their young cuckoo chicks. From the evolutionary point of view,
cuckoos aim to increase the probability of survival and reduce the probability of abandoning eggs by the host birds\cite{payne05}.
The behaviour of cuckoos is mimicked successfully in the cuckoo search algorithm, in combination
with L\'{e}vy flights to effectively search better and optimal survival strategy.
L\'{e}vy flights~\cite{brown07}, named by the French mathematician Paul L\'evy, represent a
model of random walks characterized by their step lengths which obey a power-law distribution.
Several scientific studies have shown that the search for preys by hunters follows typically the
same characteristics of L\'{e}vy flights. This model is commonly presented by small random steps followed occasionally by large jumps~\cite{brown07,shle95}.

Inspired by these behaviours and strategies, the Cuckoo Search (CS) algorithm was
developed by Xin-She Yang and Suash Deb in 2009, which was initially designed for solving multimodal functions.
It has been shown that CS can be very efficient in dealing with highly nonlinear optimization problems~\cite{yangbook2013,yangmocs2013,yang2010eng,mohamad2013proc,mohamad2013cuckoo}.

CS is summarized as the following ideal rules:
(1) each cuckoo lays one egg at a time and selects a nest randomly; (2) the best nest with the highest quality of egg
can pass onto the new generations; (3) the number of host nests is fixed, and the egg laid by
a cuckoo can be discovered by the host bird with a probability $p_a\in [0, 1]$.

CS uses a balanced combination of a local random walk and the global
explorative random walk, controlled by
a switching parameter $p_a$. The local random walk can be written as
\begin{equation}
\label{gen1}
x_i^{t+1}=x_i^t +\alpha s \otimes H(p_a-\epsilon) \otimes (x_j^t-x_k^t), \label{CS-equ1}
\end{equation}
where $x_j^t$ and $x_k^t$ are two different solutions selected randomly by random permutation,
$H(u)$ is a Heaviside function, $\epsilon$ is a random number drawn from a uniform distribution, and
$s$ is the step size. On the other hand, the global random walk is carried out using L\'evy flights
\begin{equation}
\label{gen2}
 x_i^{t+1}=x_i^t+\alpha L(s,\lambda), \label{CS-equ2}
 \end{equation}
where \begin{equation} \label{levy}
L(s,\lambda)=\frac{\lambda \Gamma(\lambda) \sin (\pi \lambda/2)}{\pi}
\frac{1}{s^{1+\lambda}}, \quad (s \gg s_0>0).
\end{equation}
Here $\alpha>0$ is the step size scaling factor, which should be related to the scales of the problem of
interest. L\'evy flights have an infinite variance with an infinite mean~\cite{yang09}.

In this approach we have taken as a basis an improved version of CS~\cite{ouaarab2014improved}. This improvement considers a new category of cuckoos that can engage a kind of surveillance on nests likely to be a host. These cuckoos use mechanisms before and after brooding such as the
observation of the host nest to decide if the chosen nest is the best
choice or not. So, from the current solution, this portion $p_c$ of cuckoos searches in the same
area a new better solution via L\'evy flights.

The goal of the improvement is to strengthen intensive search around the current solutions, using the new fraction $p_c$. The process of this fraction can be introduced in the standard algorithm of CS as shown in Algorithm~\ref{algo:RKCS}.

\begin{algorithm}[h]
\caption{Improved CS Algorithm}
\label{algo:RKCS}
\begin{algorithmic}[1]
\STATE Objective function $f(x), x=(x_1,\dots,x_m)^T$
\STATE Generate initial population of $n$ host nests $x_i(i=1,\dots,n)$
\WHILE {($t<$MaxGen) or (stop criterion)}
	\STATE \textbf{Start searching with a fraction ($p_c$) of smart cuckoos}
	\STATE Get a cuckoo randomly by L\'{e}vy flights
		\STATE Evaluate its quality/fitness $F_i$
	\STATE Choose a nest among $n$ (say, $j$) randomly
	\IF  {($F_i>F_j$)}
   		\STATE replace $j$ by the new solution;
	\ENDIF
	\STATE A fraction $(p_a)$ of worse nests are abandoned and new ones are built;
	\STATE Keep the best solutions (or nests with quality solutions);
	\STATE Rank the solutions and find the current best
\ENDWHILE
\STATE Postprocess results and visualization
\end{algorithmic}
\end{algorithm}

\section{Travelling Salesman Problem}
\label{sec:3}

To simplify the statement of the travelling salesman problem,
we can assume that we have a list of $m$ cities that must be visited by a salesperson
and returning to the departure city. To calculate the best tour
in term of distance, some rules or assumptions can be used before
starting the trip. Each city on the list must be visited exactly
once, and for each pair of cities, given that the distance between any two cities is known.
This problem is commonly called as the ``The travelling salesman problem''.

The TSP can be stated formally~\cite{davendra10} as:
Let $C = \{c_1,\dots , c_m\}$ be a set of $m$ distinct cities, $E = (c_i,~c_j): i, j \in \{1,\dots , m\}$
be the edge set, and $d_{c_ic_j}$ be a cost measure associated with the edge $(c_i,~c_j) \in E$.
The objective of the TSP is
to find the minimal length of a closed tour that visits each city once. Cities $c_i\in C$ are
represented by their coordinates $(c_{i_x} , c_{i_y})$ and
$d_{c_i c_j}=\sqrt{(c_{i_x}-c_{j_x} )^2 + (c_{i_y}-c_{j_y})^2}$ is the Euclidean distance between $c_i$ and $c_j$.

A tour can be represented as a cyclic
permutation~\cite{gutin2002traveling} $\pi = (\pi(1),\pi(2),\dots,\pi(m))$ of cities from $1$ to $m$ if $\pi(i)$ is interpreted
to be the city visited in step $i$, $i= 1,\dots,m$. The cost of a permutation (tour) is
defined as:
\begin{equation}
\label{tour}
f(\pi) = \sum^{m-1}_{i=1} d_{\pi(i) \pi(i+1)} + d_{\pi(m)\pi(1)}
\end{equation}
If $d_{c_i c_j}\neq d_{c_j c_i}$ for at least one $(c_i, c_j)$, we say that it is an Asymmetric Euclidean TSP, while if $d_{c_i c_j} = d_{c_j c_i}$
 then the TSP becomes a Symmetric Euclidean TSP, which is adopted in this paper.
Given $m$ as the number of cities to visit in the list, the total number of possible
tours covering all cities can be seen as a set of feasible solutions of the TSP and is
given as $m!$.

The challenges of solving a TSP have motivated many researchers to design various algorithms.
An effective search should be able to detect the best solution for the majority of its instances in a reasonable runtime.
The good thing about TSP is that the statement is simple and requires no mathematical background to understand,
though it is difficult to produce a good solution. However, TSP is crucially important in both
academia and applications.  Lenstra et. al. and Reinelt~\cite{lenstra75,reinelt94} gave some reviews of
direct and indirect applications of TSP in several industrial and technological fields,
such as drilling problem of printed circuit boards (PCBs), overhauling gas turbine
engines, X-ray crystallography, computer wiring, order-picking problem in warehouses, vehicle routing, and mask plotting in PCB production.

\section{Random-Key Encoding Scheme}
\label{sec:4}
Random-key encoding scheme~\cite{bean1994genetic} is a technique that can be used to
transform a position in a continuous space and convert it into a combinatorial one.
It uses a vector of real numbers by associating each number with a weight. These weights are used to generate one combination as a solution.

The random real numbers drawn uniformly from $[0,1)$ compose a vector showed in Fig.~\ref{fig:RKES}. On the other hand, the  combinatorial vector is
composed of integers ordered according to the weights of real numbers in the first vector, illustrated as follows (Fig.~\ref{fig:RKES}):
\begin{figure}[h]
\begin{tikzpicture}
\tikzstyle{every path}=[very thick]

\edef\sizetape{0.68cm}
\tikzstyle{tmtape}=[draw,minimum size=\sizetape, line width=0.1]

\begin{scope}[start chain=1 going right,node distance=-0.1mm]
    \node [on chain=1,tmtape,draw=none] (a) {{\small Random keys:}};
    \node [on chain=1,tmtape] (b) {0.8};
    \node [on chain=1,tmtape] (c) {0.5};
    \node [on chain=1,tmtape] (d) {0.7};
    \node [on chain=1,tmtape] (e) {0.1};
    \node [on chain=1,tmtape] (f) {0.4};
    \node [on chain=1,tmtape] (g) {0.2};
\end{scope}
    \node [below of=a,tmtape,draw=none] {{\small Decoded as:}};
    \node [below of=b,tmtape] {6};
    \node [below of=c,tmtape] {4};
    \node [below of=d,tmtape] {5};
    \node [below of=e,tmtape] {1};
    \node [below of=f,tmtape] {3};
    \node [below of=g,tmtape] {2};
\label{fig:RKES}
\end{tikzpicture}
\caption{Random key encoding scheme.}
\end{figure}
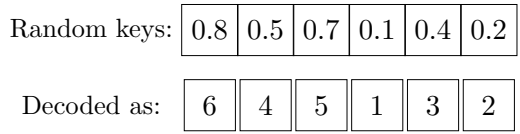

\section{Random Key CS for TSP}
\label{sec:5}

Random keys are an encoding scheme which was used early with genetic algorithms for sequencing
and optimization problems by Bean~\cite{bean1994genetic}. It is based on random real numbers in a continuous space to encode solutions in a combinatorial space. These random
numbers, presented as tags, are generated from $[0.1)^{m}$ space, where $m$ is the size of the TSP instance to be solved (or the dimension of this space).
Our approach here extends this idea by performing L\'evy flights distribution to generate the random numbers.
This allows an improved way to balance the search for solutions in local areas as well as global areas. In this work, we will thus use
both random walks and L\'evy flights~\cite{yang2011nature} whose step lengths are chosen from a probability distribution with a power-law tail.

\subsection{Solution Representation}
\label{sec:5.1}

\begin{figure*}
\centerline{\epsfig{file=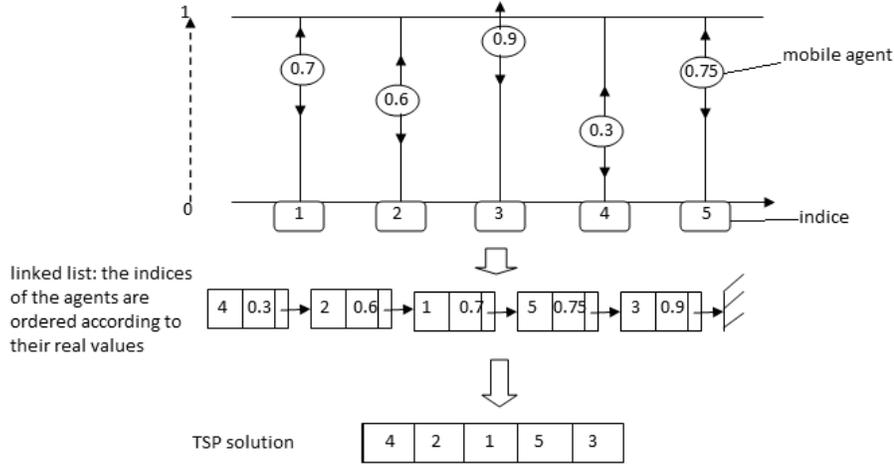,height=7cm}}
\caption{Procedure of random-keys to generate a TSP solution.}
\label{fig:init}
\end{figure*}

Figure~\ref{fig:init} and Algorithm~\ref{algo:init} present the steps of generating a TSP solution using random keys. First, agents
are randomly positioned according to their real values in $[0,1)$. Each agent has an integer index regardless
of his ascending order among other agents in the linked list (see Fig.~\ref{fig:init}). All agents are ordered according to their weights (real numbers) and their indices
form together an initial solution of the TSP instance. So, this essentially means that the integers/agent indices, here, correspond to
 the city index and the order of agents is the visiting order of the cities.

\begin{algorithm}
\caption{Initial Solution Algorithm}
\label{algo:init}
\begin{algorithmic}[1]
\STATE Set of $m$ agents $a_i \; (i=1,\dots,m)$;
\FOR { $i=1$ \TO $m$ }
	\STATE Assign a random real number in $[0.1)$ for agent $a_i$;
\ENDFOR
\FOR { $i=1$ \TO $m$ }
	\STATE Get the order of agent $a_i$ according to his weight;
\ENDFOR
\FOR { $i=1$ \TO $m$ }
	\STATE The agent index is the city index;
	\STATE The agent order is the city visiting order;
\ENDFOR
\STATE Return a solution of cities' visiting order and weights;
\end{algorithmic}
\end{algorithm}

\subsection{Displacement}
\label{sec:5.2}

\begin{figure*}
\centerline{\epsfig{file=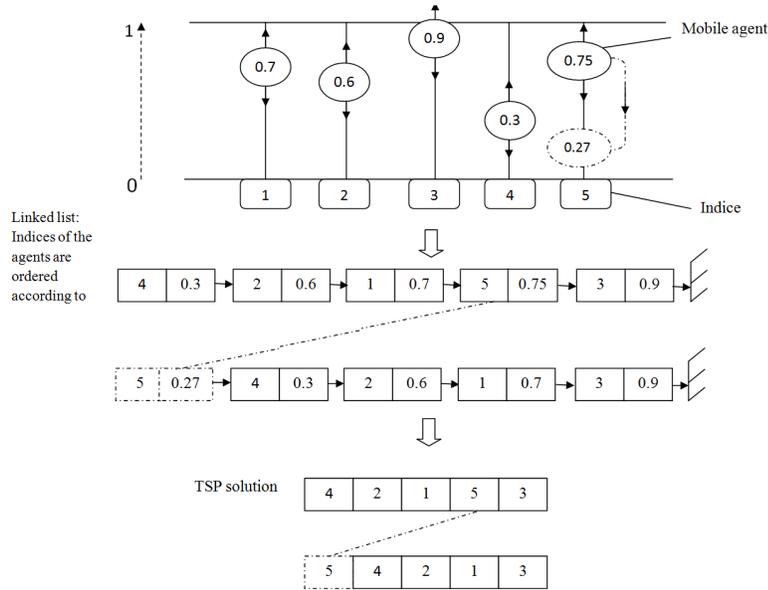,height=8cm}}
\caption{Procedure of random keys move to generate new TSP solution}
\label{fig:move}
\end{figure*}

The procedure of generating new solutions by a perturbation in the real space can lead to
some issues  when these agents start to move in $[0,1)$. Such moves can affect
the order of agents in the linked list and therefore a new TSP solution can be generated
as shown in Fig.~\ref{fig:move}. The displacement of each agent is guided directly by L\'evy flights.
The order of each city is changed by small perturbations or big jumps according to the
values generated in their weights via L\'evy flights.

Our approach here is mainly based on two types of search moves: 1) global search carried out
on solution areas guided by the movements (following L\'evy flights) of agents; 2) local search
which detects the best solutions in the areas found by the agents. The combination of both
local and global search moves can improve the performance. Briefly, RKCS begins with a search for
new promising areas. It combines intensification and diversification via small steps and large jumps
to distant areas. When pointing on an area found by L\'evy flights,
the best solution in this area is detected and another search is triggered to generate a new one
via L\'evy flights. So, we can summarize these steps by the following Algorithm~\ref{algo:move}:

\begin{algorithm}
\caption{Generating New Solution}
\label{algo:move}
\begin{algorithmic}[1]
\STATE Solution $S$ of $m$ cities/agents $a_i \; (i=1,\dots,m)$;
\STATE Select randomly $l$ agents ($1\leq l \leq m$).
\FOR { $i=1$ \TO $l$ }
	\STATE Assign new position via L\'evy flights (Equation~\ref{CS-equ2}) for agent $a_i$;
	\STATE Reposition $a_i$ in the linked list (see Figure~\ref{algo:move});
	\STATE Update $S$;
\ENDFOR
\STATE Return the new generated solution $S$;
\end{algorithmic}
\end{algorithm}

\subsection{Local Search}
\label{sec:5.3}
\begin{figure}
\centerline{\epsfig{file=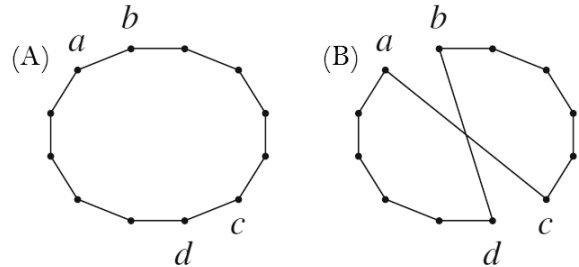,height=4cm}}
\caption{2-opt move, (A) Initial
tour. (B) The tour created by
2-opt move [the edges (a, b)
and (c, d) are removed, while
the edges (a, c) and (b, d) are added.}
\label{fig:opt}
\end{figure}

After finding a new area by L\'evy flights, a local search is performed to detect the best solution in this area.
For this local search, 2-opt move~\cite{croes58} is used where it removes two edges in the TSP solution and reconnects the new two created paths, in a different possible way as showed in Fig.~\ref{fig:opt}. In the minimization case, it is done only if the new tour is shorter than the current one.
Obviously, this process is repeated until no further improvement is possible or when a given number of steps is reached.

\begin{algorithm}
\caption{Steepest Descent Algorithm}
\label{algo:descent}
\begin{algorithmic}[1]
\STATE Objective function $f(x), x=(x_1,\dots,x_m)^T$;
\STATE Initial TSP solution $S_0$ of $m$ ordered cities;
\STATE Current solution $S,~S\longleftarrow S_0$;
\STATE Choose $stop$ boolean value, $stop\longleftarrow FALSE$;
\WHILE {$stop=FALSE$}
	\STATE Choose the best neighbour $S_v$ of $S$ via $2$-$Opt$ moves;
	\IF {$f(S_v) < f(S)$}
		\STATE Replace $S$ by $S_v$;
	\ELSE
		\STATE $stop\longleftarrow TRUE$
	\ENDIF
\ENDWHILE
\STATE Return $S$;
\end{algorithmic}
\end{algorithm}

Steepest Descent (Algorithm~\ref{algo:descent}) is a simple local search method that can be easily trapped in a local minimum, and generally, it can not find a good quality solutions. We choose this simplified local search ``Steepest Descent'' method to show the performance of CS combined with RK. It allowed us to generate solutions of good quality, without introducing an advanced local search method.

\subsection{RKCS Algorithm}
\label{sec:5.4}
Using the same steps of Improved CS~\cite{ouaarab2014improved} and as summarized in Algorithm~\ref{algo:RKCS}, before starting the search process, RKCS generates the random initial solution or population as explained in Fig.~\ref{fig:init} and Algorithm~\ref{algo:init}. Generating a random initial population is to show how RKCS can find good solutions in the search space without using an enhancement pre-processes.

The second phase is triggering the $p_c$ portion of smart cuckoos. These cuckoos begin by exploring new areas from the current solutions. As shown in Fig.~\ref{fig:move}, they use L\'evy flights to move in the real space and interpreting this move to have a new TSP solution in the new area. The second step is to find a good solution in this area following Algorithm~\ref{algo:descent}.

After $p_c$ portion phase, RKCS employs one cuckoo to search for a new good solution, starting from the best solution of the population. It proceeds, like the second phase ($p_c$ portion phase), in two steps. Firstly, it locates a new area, from the best solution, via L\'evy flights and then finds a good solution in this area. The found solution is compared with a random selected solution in the population. The best one of the both solutions earns its place in the population.

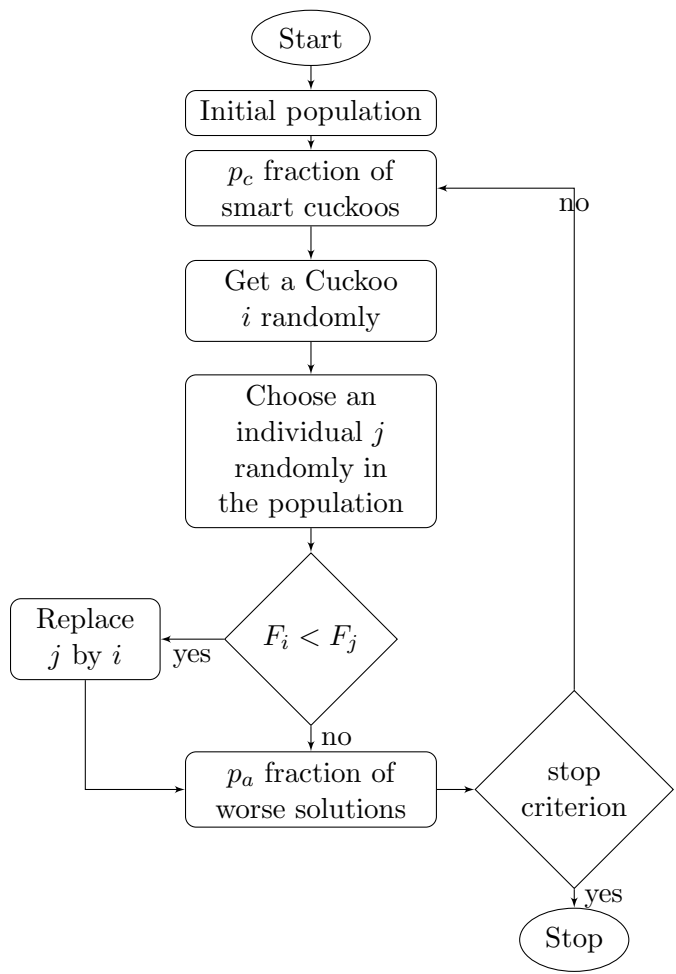
\begin{figure}
\begin{tikzpicture}[node distance = 1cm, auto]
    \node [cloud] (init) {Start};
    \node [block, below of=init] (pop) {Initial population};
    \node [block, below of=pop] (pc) {$p_c$ fraction of smart cuckoos};
    \node [block, below of=pc, node distance = 1.5cm] (cuckoo) {Get a Cuckoo $i$ randomly};
    \node [block, below of=cuckoo, node distance = 2cm] (indiv) {Choose an individual $j$ randomly in the population};
    \node [decision, below of=indiv, node distance = 2.5cm] (decide) {$F_i<F_j$};
    \node [block, left of=decide, node distance = 3cm, text width=4.5em] (update) {Replace $j$ by $i$};
    \node [block, below of=decide, node distance = 2cm] (pa) {$p_a$ fraction of worse solutions};
    \node [decision, right of=pa, node distance = 3.5cm] (max) {stop criterion};
    \node [cloud, below of=max, node distance = 2cm] (stop) {Stop};
    \path [line] (init) -- (pop);
    \path [line] (pop) -- (pc);
    \path [line] (pc) -- (cuckoo);
    \path [line] (cuckoo) -- (indiv);
    \path [line] (indiv) -- (decide);
    \path [line] (decide) -- node {yes}(update);
    \path [line] (decide) -- node {no} (pa);
    \path [line] (update) |- (pa);
    \path [line] (pa) -- (max);
    \path [line] (max) |- node {no} (pc);
    \path [line] (max) -- node {yes} (stop);
\end{tikzpicture}
\caption{The flowchart of the RKCS algorithm.}
\label{fig:flowchart}
\end{figure}

The last phase is for the worst and abandoned solutions that will be replaced by new ones. They start searching, for a new good solution, far from the best solution in the population by a big jump. In this case a big jump is perturbing more agents via L\'evy flights.  All these phases are illustrated in the flowchart of RKCS~(Fig.~\ref{fig:flowchart}).

\section{Experimental results}
\label{sec:6}

We have implemented the proposed random key cuckoo search and tested it using the well-known TSPLIB library~\cite{reinelt95}.
For each instance, 30 independent runs have been carried out. The properly selected parameter values used for the experiments of RKCS algorithm are shown in Table~\ref{tab:para}.

\setlength{\tabcolsep}{4pt}
\begin{table}[h]
\begin{center}
\caption{Parameter settings for RKCS algorithm.}
\label{tab:para}
\begin{tabular}{llll}
\hline\noalign{\smallskip}
Parameter & Value & Meaning \\
\noalign{\smallskip}
\hline
\noalign{\smallskip}
$n$  & $30$ & Population size	\\
$p_c$ & $0.6$ & Portion of smart cuckoos \\
$p_a$ & $0.2$ & Portion of bad solutions\\
$MaxGen$ & $500$ & Maximum number of iterations\\
$\alpha$ & $0.01$ & Step size\\
$\lambda$ & $1$ & Index \\
\hline
\end{tabular}
\end{center}
\end{table}
\setlength{\tabcolsep}{1.4pt}

Table~\ref{tab:reslt} shows the test results of running RKCS algorithm to solve some benchmark instances of the symmetric TSP from
the TSPLIB library~\cite{reinelt95}. The first column corresponds to the name of instances with their optimum in parentheses.
The column `Best' shows the length of the best solution found by RKCS, the column `Average'
gives the average solution found by RKCS, the column `Worst' shows the length of the worst
solution length among the 30 independent runs of the RKCS algorithm.

 These results confirm that the proposed approach is able to find good or the optimum solution for the tested instances (`bold' in the Table~\ref{tab:reslt} shows that RKCS reaches the optimal solution of the tested instance). Therefor, we can say that the random-key encoding scheme can be a very useful tool for switching from
continuous to combinatorial spaces. It allows operators of the continuous space to behave freely, then projecting the changes made ​​by these operators in the combinatorial space.
It also facilitates a better control in balancing intensification and diversification through L\'evy flights, which make intensified small steps in a limited region followed by a big explorative jump to a distant region. Using the real numbers, L\'evy flights can easily act with the notion of distance and can define clearly small or big steps. Then RK projects these changes in the space of TSP solutions.

\setlength{\tabcolsep}{4pt}
\begin{table}[h]
\begin{center}
\caption{Results of random-key cuckoo search for the travelling salesman problem}
\label{tab:reslt}
\begin{tabular}{llll}
\hline\noalign{\smallskip}
Instance(opt) & Best & Average & Worst\\
\noalign{\smallskip}
\hline
\noalign{\smallskip}
eil51(426)  & \textbf{426} & 426.9 & 430\\
berlin52(7542) & \textbf{7542} & 7542 & 7542\\
st70(675) & \textbf{675} & 677.3 & 684\\
pr76(108159) & \textbf{108159} & 108202 & 109085\\
eil76(538) & \textbf{538} & 539.1 & 541 \\
kroA100(21282) & \textbf{21282} & 21289.65 & 21343\\
eil101(629) & \textbf{629} & 631.1 & 636\\
bier127(118282) & \textbf{118282} & 118798.1 & 120773\\
pr136(96772) & 97046 & 97708.9 & 98936\\
pr144(58537) & \textbf{58537} & 58554.45 & 58607\\
ch130(6110) & 6126 & 6163.3 & 6210\\
\hline
\end{tabular}
\end{center}
\end{table}
\setlength{\tabcolsep}{1.4pt}

\setlength{\tabcolsep}{4pt}
\begin{table}
\begin{center}
\caption{Comparison of experimental results of RKCS with all algorithms cited in~\cite{chen11}}
\label{tab:comp}
\begin{tabular}{lllllll}
\hline\noalign{\smallskip}
 Instances & eil51 & st70& rd100& pr124& rat195& \\
\noalign{\smallskip}
\hline
Best & 426& 675& 7910& 59030& 2323& Average\\
\hline
\noalign{\smallskip}
GA  & 2.58\% & 2.35\% & 5.27\%& 2.74\%& 6.68\%& 3.92\%\\
ACO & 1.08\% & 1.98\% & 3.14\%& 1.23\%& 2.59\%& 2.00\%\\
PSO & 1.12\% & 2.32\% & 2.65\%& 1.98\%& 3.45\%& 2.30\%\\
AIS & 1.22\% & 1.79\% & 2.03\%& 1.45\%& 2.77\%& 1.85\%\\
IWDA & 4.08\% & 5.20\% & 4.97\%& 6.12\%& 5.34\%& 5.14\% \\
BCO & 2.19\% & 3.01\% & 2.44\%& 2.78\%& 3.43\%& 2.77\%\\
EM & 2.67\% & 3.05\% & 2.78\%& 3.45\%& 5.45\%& 3.48\%\\
Hr-GSA & 0.54\% & 0.34\% & 1.12\%& 1.05\%& 2.56\%& 1.12\%\\
RKCS & \textbf{0.0}\% & \textbf{0.0}\% & \textbf{0.0}\% & \textbf{0.0}\%& 0.38 & 0.07  \\
\hline
\end{tabular}
\end{center}
\end{table}
\setlength{\tabcolsep}{1.4pt}

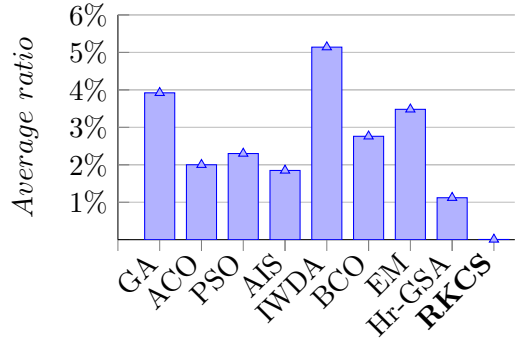
\begin{figure}[h]
    \centering
    \begin{tikzpicture}
        \begin{axis}[
            ybar,
            xmin = 0,
            xmax = 4.75,
            ymin = 0,
            ymax = 15,
            axis x line* = bottom,
            axis y line* = left,
            ylabel= $Average~ratio$,
            width= 0.45\textwidth,
            height = 0.3\textwidth,
            ymajorgrids = true,
            bar width = 4mm,
            xticklabels = \empty,
            yticklabels = \empty,
            extra x ticks = {0.5,1,1.5,2,2.5,3,3.5,4,4.5},
            extra x tick labels = {GA, ACO, PSO, AIS, IWDA, BCO, EM, Hr-GSA, \textbf{RKCS}},
            x tick label style={rotate=45,anchor=east},
            extra y ticks = {2.5,5,7.5,10,12.5,15},
            extra y tick labels = {1\%, 2\%, 3\%, 4\%, 5\%, 6\%},
            ]
            \addplot+ plot[mark=triangle*] coordinates {
                (0.5,9.8)
                (1,5)
                (1.5,5.75)
                (2,4.62)
                (2.5,12.85)
                (3,6.9)
                (3.5,8.7)
                (4,2.8)
                (4.5,0.015)
            };
        \end{axis}
    \end{tikzpicture}
    \caption[Average ratio of the solutions found to the optimal solution, for instances eil51, st70, rd100, pr124 and rat195]{Average ratio of the solutions found to the optimal solution, for instances eil51, st70, rd100, pr124 and rat195}
    \label{fig:new_ev}
\end{figure}

The RKCS experimental results  are then compared with all other algorithms tested in `Hybrid Gravitational Search Algorithm with Random-key Encoding
Scheme Combined with Simulated Annealing'~\cite{chen11}. Table~\ref{tab:comp} and Fig.~\ref{fig:new_ev} show that RKCS outperforms all these algorithms in the tested instances. A good balance between exploration and exploitation of space areas and a simple local search technique are significant components to reach good results. This confirms that the proposed random-key encoding scheme can provide a good performance by combining
 global and local search strength in one entity within RKCS via L\'evy flights.

 In Table~\ref{tab:comp}, the `Best' denotes the best known-so-far optimal solution quality, and the other results recorded
 were the ratio of the solutions found by each algorithm to the optimal solution over 30 independent runs.

\section{Conclusion}

In this proposed approach, we used the random-key encoding scheme combined
with the cuckoo search to develop the random-key cuckoo search (RKCS) algorithm for solving TSP.
By testing the standard TSP instances' library, we can confirm that the proposed approach
is very efficient and can obtain good results.

Random keys have been used to switch between the continuous and the combinatorial search space,
which enable cuckoo search to provide a good search mechanism with a fine balance
between intensification and diversification.

However, RKCS can be altered or improved by many ways. For example, it can be useful to adapt the agent moves with some
existing or new move operators in the combinatorial space. In addition, the tuning of algorithm-dependent
parameters can be also fruitful~\cite{yangsta2013}, which may provide further research topics
to see if the proposed approach can be further improved.

It can be expected that our approach can be used to
solve other combinatorial optimization problems such as routing, scheduling and even mixed-integer programming problems. We can also generalize this work to solve some kinds of TSP problems such as Asymmetric~\cite{cirasella2001asymmetric} , Spherical~\cite{ouyang2013novel} and generalized~\cite{snyder2006random} TSP.

\bibliographystyle{spmpsci}

\end{document}